\documentclass[11pt]{article}
\usepackage{NRC-47167}
\usepackage{times}
\usepackage{latexsym}

\title{Word Sense Disambiguation by Web Mining \\
  for Word Co-occurrence Probabilities}

\author{Peter D. TURNEY \\
  Institute for Information Technology \\
  National Research Council of Canada \\
  Ottawa, Ontario, Canada, K1A 0R6 \\
  peter.turney@nrc-cnrc.gc.ca}

\date{}

\begin{document}

\lefthyphenmin=3 
\righthyphenmin=2

\maketitle

\begin{abstract}
This paper describes the National Research Council (NRC)
Word Sense Disambiguation (WSD) system, as applied to the
English Lexical Sample (ELS) task in Senseval-3. The NRC system 
approaches WSD as a classical supervised machine learning problem,
using familiar tools such as the Weka machine learning software 
and Brill's rule-based part-of-speech tagger. Head words are
represented as feature vectors with several hundred features.
Approximately half of the features are syntactic and the other
half are semantic. The main novelty in the system is the method for
generating the semantic features, based on word \hbox{co-occurrence} 
probabilities. The probabilities are estimated using 
the Waterloo MultiText System with a corpus of about one terabyte of 
unlabeled text, collected by a web crawler.
\end{abstract}

\section{Introduction}

The \hbox{Senseval-3} English Lexical Sample (ELS) task requires
disambiguating 57 words, with an average of roughly 140 
training examples and 70 testing examples of each word. Each example
is about a paragraph of text, in which the word that is
to be disambiguated is marked as the {\em head} word. The average
head word has around six senses. The training
examples are manually classified according to the intended sense of the head
word, inferred from the surrounding context. The task is to 
use the training data and any other relevant information to
automatically assign classes to the testing examples.

This paper presents the National Research Council (NRC)
Word Sense Disambiguation (WSD) system, which generated our four entries for 
the Senseval-3 ELS task (NRC-Fine, NRC-Fine2, NRC-Coarse, and NRC-Coarse2).
Our approach to the ELS task is to treat it as a classical supervised 
machine learning problem. Each example is represented as a feature
vector with several hundred features. Each of the 57 ambiguous words
is represented with a different set of features. Typically, around half of the features are 
syntactic and the other half are semantic. After the raw examples are
converted to feature vectors, the Weka machine learning software
is used to induce a model of the training data and predict the
classes of the testing examples \cite{Witten:99}.

The syntactic features are based on part-of-speech tags,
assigned by a rule-based tagger \cite{Brill:94}. The main innovation
of the NRC WSD system is the method for generating the semantic features, which are derived
from word \hbox{co-occurrence} probabilities. We estimated these probabilities
using the Waterloo MultiText System with a corpus of about one terabyte of 
unlabeled text, collected by a web crawler \cite{Clarke:95,Clarke:00,Terra:03}.

In Section~\ref{sec:description}, we describe the NRC WSD system. 
Our experimental results are presented in Section~\ref{sec:results} and
we conclude in Section~\ref{sec:conclusion}.

\section{System Description}\label{sec:description}

This section presents various aspects of the system in roughly the order
in which they are executed. The following definitions will simplify
the description.

\noindent \textbf{Head Word:} One of the 57 words that are to be disambiguated.

\noindent \textbf{Example:} One or more contiguous sentences, illustrating the 
usage of a head word. 

\noindent \textbf{Context:} The non-head words in an example.

\noindent \textbf{Feature:} A property of a head word in a context. For instance, 
the feature \texttt{tag\_hp1\_NNP} is the property of having (or not having) 
a proper noun (\texttt{NNP} is the part-of-speech tag for a proper noun) immediately 
following the head word (\texttt{hp1} represents the location {\em head plus one}).

\noindent \textbf{Feature Value:} Features have values, which depend on the
specific example. For instance, \texttt{tag\_hp1\_NNP}
is a binary feature that has the value 1 ({\em true}: the following word {\em is} a
proper noun) or 0 ({\em false}: the following word is {\em not} a proper noun).

\noindent \textbf{Feature Vector:} Each example is represented by a vector.
Features are the dimensions of the vector space and a vector of feature values
specifies a point in the feature space. 

\subsection{Preprocessing}

The NRC WSD system first assigns part-of-speech tags to the
words in a given example \cite{Brill:94}, and then extracts
a nine-word window of tagged text, centered on the head word
(i.e., four words before and after the head word).
Any remaining words in the example are ignored (usually 
most of the example is ignored).
The window is not allowed to cross sentence boundaries.
If the head word appears near the beginning or end of the sentence,
where the window may overlap with adjacent sentences, special
{\em null} characters fill the positions of any missing words in the window.

In rare cases, a head word appears more than once in an example.
In such cases, the system selects a single window, giving preference
to the earliest occurring window with the least nulls. Thus each 
example is converted into one nine-word window of tagged text. 
Windows from the training examples for a given 
head word are then used to build the feature set for that head word. 

\subsection{Syntactic Features}

Each head word has a unique set of feature names, describing how 
the feature values are calculated.

\noindent \textbf{Feature Names:} Every syntactic feature has a name of the form 
{\em matchtype}\_{\em position}\_{\em model}.
There are three {\em matchtypes}, \texttt{ptag}, \texttt{tag}, and
\texttt{word}, in order of increasingly strict matching. A \texttt{ptag} match
is a {\em partial tag match}, which counts similar part-of-speech tags, such as
\texttt{NN} (singular noun), \texttt{NNS} (plural noun), \texttt{NNP} (singular proper noun),
and \texttt{NNPS} (plural proper noun), as equivalent. A \texttt{tag} match
requires exact matching in the part-of-speech tags for the word and the model.
A \texttt{word} match requires that the word and the model are exactly the same,
letter-for-letter, including upper and lower case.

There are five {\em positions}, \texttt{hm2} (head minus two), \texttt{hm1} 
(head minus one), \texttt{hd0} (head), \texttt{hp1} (head plus one), and
\texttt{hp2} (head plus two). Thus syntactic features use only a five-word
sub-window of the nine-word window.

The syntactic feature names for a head word are generated by all of the
possible legal combinations of {\em matchtype}, {\em position}, and {\em model}.
For \texttt{ptag} names, the {\em model} can be any partial tag.
For \texttt{tag} names, the {\em model} can be any tag. For \texttt{word} names, 
the {\em model} names are not predetermined; they are extracted from the training 
windows for the given head word. For instance, if a training window
contains the head word followed by ``of'', then one of the features will
be \texttt{word\_hp1\_of}. 

For \texttt{word} names, the {\em model} names are not allowed to be words
that are tagged as nouns, verbs, or adjectives. These words are reserved for
use in building the semantic features.

\noindent \textbf{Feature Values:} The syntactic features are all binary-valued. Given a feature 
with a name of the form {\em matchtype}\_{\em position}\_{\em model}, the feature value 
for a given window depends on whether there is a match of {\em matchtype}
between the word in the position {\em position} and the model {\em model}.
For instance, the value of \texttt{tag\_hp1\_NNP} depends on whether the given window
has a word in the position \texttt{hp1} (head plus one) with a
\texttt{tag} (part-of-speech tag) that matches \texttt{NNP} (proper noun).
Similarly, the feature \texttt{word\_hp1\_of} has the
value 1 ({\em true}) if the given window contains the head word followed by ``of'';
otherwise, it has the value 0 ({\em false}). 

\subsection{Semantic Features}

Each head word has a unique set of feature names, describing how 
the feature values are calculated.

\noindent \textbf{Feature Names:} Most of the semantic features have names 
of the form {\em position}\_{\em model}. 
The {\em position} names can be \texttt{pre} (preceding) or \texttt{fol} (following).
They refer to the nearest noun, verb, or adjective that precedes or follows
the head word in the nine-word window. 

The {\em model} names are extracted from the training windows for the head
word. For instance, if a training window contains the word ``compelling'', 
and this word is the nearest noun, verb, or adjective that 
precedes the head word, then one of the features will be \texttt{pre\_compelling}.

A few of the semantic features have a different form of name, 
\texttt{avg}\_{\em position}\_{\em sense}. In names of this form,
{\em position} can be \texttt{pre} (preceding) or \texttt{fol} (following),
and {\em sense} can be any of the possible senses (i.e., classes, labels) of the head word.

\noindent \textbf{Feature Values:} The semantic features are all real-valued. 
For feature names of the form {\em position}\_{\em model},
the feature value depends on the semantic similarity
between the word in position {\em position} and the
model word {\em model}. 

The semantic similarity between two words is estimated by their
Pointwise Mutual Information, ${\rm PMI}(w_1,w_2)$, using Information 
Retrieval \cite{Turney:01,Terra:03}:

\[
{\rm PMI}(w_1 ,w_2 ) = {\rm log}_2 \left( {\frac{{p(w_1 \wedge w_2 )}}{{p(w_1 )p(w_2 )}}} \right).
\]

\def\tablesize{\fontsize{10}{11} \selectfont}

\begin{table*}[t] \centering 
\begin{tablesize}
\begin{tabular}{l}
\hline
weka.classifiers.meta.Bagging \\
~~~~-W weka.classifiers.meta.MultiClassClassifier  \\
~~~~~~~~-W weka.classifiers.meta.Vote \\
~~~~~~~~~~~~-B weka.classifiers.functions.supportVector.SMO \\
~~~~~~~~~~~~-B weka.classifiers.meta.LogitBoost -W weka.classifiers.trees.DecisionStump \\
~~~~~~~~~~~~-B weka.classifiers.meta.LogitBoost -W weka.classifiers.functions.SimpleLinearRegression \\
~~~~~~~~~~~~-B weka.classifiers.trees.adtree.ADTree \\
~~~~~~~~~~~~-B weka.classifiers.rules.JRip \\
\hline
\end{tabular}
\vspace{-5pt}
\caption[]{Weka (version 3.4) commands for processing the feature vectors.}\label{weka}
\end{tablesize}
\end{table*}

\noindent We estimate the probabilities in this equation by issuing
queries to the Waterloo MultiText System \cite{Clarke:95,Clarke:00,Terra:03}.
Laplace smoothing is applied to the PMI estimates, to avoid division by zero.

${\rm PMI}(w_1,w_2)$ has a value of zero when the two words are statistically 
independent. A high positive value indicates that the two words tend to
co-occur, and hence are likely to be semantically related. A negative value
indicates that the presence of one of the words suggests the absence of the
other. Past work demonstrates that PMI is a good estimator of semantic
similarity \cite{Turney:01,Terra:03} and that features based on PMI
can be useful for supervised learning \cite{Turney:03}.

The Waterloo MultiText System allows us to set the neighbourhood size
for co-occurrence (i.e., the meaning of $w_1 \wedge w_2$). 
In preliminary experiments with the ELS data from
Senseval-2, we got good results with a neighbourhood size of 20 words.

For instance, if $w$ is the noun, verb, or adjective that precedes the head 
word and is nearest to the head word in a given window, then the value of \texttt{pre\_compelling} is 
${\rm PMI}(w,{\rm compelling)}$. If there is no preceding 
noun, verb, or adjective within the window, the value is set to zero.

In names of the form \texttt{avg}\_{\em position}\_{\em sense},
the feature value is the average of the feature values of the corresponding
features. For instance, the value of \texttt{avg\_pre\_argument\_1\_10\_02} is
the average of the values of all of the \texttt{pre}\_{\em model} features,
such that {\em model} was extracted from a training window in which the head word was labeled
with the sense \texttt{argument\_1\_10\_02}.

The idea here is that, if a testing example should be labeled, say, \texttt{argument\_1\_10\_02},
and $w_1$ is a noun, verb, or adjective that is close to the head word in the testing
example, then ${\rm PMI}(w_1,w_2)$ should be relatively high when $w_2$ is extracted
from a training window with the same sense, \texttt{argument\_1\_10\_02}, but
relatively low when $w_2$ is extracted from a training window with a different sense.
Thus \texttt{avg}\_{\em position}\_\texttt{argument\_1\_10\_02} is likely to be relatively
high, compared to other \texttt{avg}\_{\em position}\_{\em sense} features.

All semantic features with names of the form {\em position}\_{\em model} are normalized 
by converting them to percentiles. The percentiles are calculated separately for each
feature vector; that is, each feature vector is normalized internally, with respect to
its own values, not externally, with respect to the other feature vectors. The \texttt{pre}
features are normalized independently from the \texttt{fol} features. The semantic
features with names of the form \texttt{avg}\_{\em position}\_{\em sense} are
calculated after the other features are normalized, so they do not need any further
normalization. Preliminary experiments with the ELS data from Senseval-2 supported
the merit of percentile normalization, which was also found useful in another
application where features based on PMI were used for supervised learning \cite{Turney:03}.

\subsection{Weka Configuration}

Table~\ref{weka} shows the commands that were used to execute Weka \cite{Witten:99}.
The default parameters were used for all of the classifiers. Five base classifiers
(-B) were combined by voting. Multiple classes were
handled by treating them as multiple two-class problems, using a 1-against-all strategy. 
Finally, the variance of the system was reduced with bagging. 

We designed the Weka configuration by evaluating many different Weka
base classifiers on the Senseval-2 ELS data, until we had identified five
good base classifiers. We then experimented with combining the base
classifiers, using a variety of meta-learning algorithms. The resulting system
is somewhat similar to the JHU system, which had the best ELS scores in Senseval-2
\cite{Yarowsky:01}. The JHU system combined four base classifiers using a form of voting, 
called Thresholded Model Voting \cite{Yarowsky:01}.

\def\tablesize{\fontsize{10}{11} \selectfont}

\begin{table*}[t] \centering 
\begin{tablesize}
\begin{tabular}{lrr}
\hline
System                    & Fine-Grained Recall & Coarse-Grained Recall \\
\hline
Best Senseval-3 System    & 72.9\%              & 79.5\%               \\
NRC-Fine                  & 69.4\%              & 75.9\%               \\
NRC-Fine2                 & 69.1\%              & 75.6\%               \\
NRC-Coarse                & NA                  & 75.8\%               \\
NRC-Coarse2               & NA                  & 75.7\%               \\
Median Senseval-3 System  & 65.1\%              & 73.7\%               \\
Most Frequent Sense       & 55.2\%              & 64.5\%               \\
\hline
\end{tabular}
\vspace{-5pt}
\caption[]{Comparison of NRC-Fine with other Senseval-3 ELS systems.}\label{recall}
\end{tablesize}
\end{table*}

\subsection{Postprocessing}

The output of Weka includes an estimate of the probability for
each prediction. When the head word is frequently labeled U (unassignable) in 
the training examples, we ignore U examples during training, and
then, after running Weka, relabel the lowest probability testing examples as U.

\section{Results}\label{sec:results}

A total of 26 teams entered 47 systems (both supervised and unsupervised) 
in the Senseval-3 ELS task. Table~\ref{recall} compares the fine-grained 
and coarse-grained scores of our four entries with other Senseval-3 systems. 

With NRC-Fine and NRC-Coarse, each semantic feature was scored by calculating its PMI
with the head word, and then low scoring semantic features were dropped. With NRC-Fine2
and NRC-Coarse2, the threshold for dropping features was changed, so that many more
features were retained. The Senseval-3 results suggest that it is better to drop more features.
 
NRC-Coarse and NRC-Coarse2 were designed to maximize the coarse score, by training
them with data in which the senses were relabeled by their coarse sense equivalence classes.
The fine scores for these two systems are meaningless and should be ignored. The Senseval-3
results indicate that there is no advantage to relabeling.

The NRC systems scored roughly midway between the best and median systems.
This performance supports the hypothesis that corpus-based semantic features can
be useful for WSD. In future work, we plan to design a system that combines
corpus-based semantic features with the most effective elements of the other
Senseval-3 systems.

For reasons of computational efficiency, we chose a relatively narrow window
of nine-words around the head word. We intend to investigate whether a larger
window would bring the system performance up to the level of the best Senseval-3
system.

\section{Conclusion}\label{sec:conclusion}

This paper has sketched the NRC WSD system for the ELS task in
Senseval-3. Due to space limitations, many details were omitted,
but it is likely that their impact on the performance is relatively small.

The system design is relatively straightforward and classical. 
The most innovative aspect of the system is the set of
semantic features, which are purely corpus-based; no lexicon was used. 

\section*{Acknowledgements}

We are very grateful to Egidio Terra, Charlie Clarke, and the
School of Computer Science of the University of Waterloo, for
giving us a copy of the Waterloo MultiText System. Thanks to
Diana Inkpen, Joel Martin, and Mario Jarmasz for helpful discussions. 
Thanks to the organizers of Senseval for their service to the
WSD research community. Thanks to Eric Brill and the developers
of Weka, for making their software available.

\bibliographystyle{NRC-47167}
\bibliography{NRC-47167}

\end{document}